\documentclass[10pt,twocolumn,letterpaper]{article}

\usepackage[pagenumbers]{wacv} 

\usepackage{graphicx}
\usepackage{amsmath}
\usepackage{amssymb}
\usepackage{booktabs}

\usepackage[pagebackref,breaklinks,colorlinks]{hyperref}

\usepackage[capitalize]{cleveref}
\crefname{section}{Sec.}{Secs.}
\Crefname{section}{Section}{Sections}
\Crefname{table}{Table}{Tables}
\crefname{table}{Tab.}{Tabs.}

\usepackage{enumitem}
\usepackage{multirow}

\newcommand{\secref}[1]{Section~\ref{sec:#1}}
\newcommand{\figref}[1]{Figure~\ref{fig:#1}} 
\newcommand{\tabref}[1]{Table~\ref{tab:#1}}

\newcommand{\defref}[1]{Definition~\ref{def:#1}}
\newcommand{\topic}[1]
{
\noindent \textbf{#1}
}
\newcommand{\tb}[1]{\textbf{#1}}

\def\wacvPaperID{764} 
\def\confName{WACV}
\def\confYear{2024}

\begin{document}

\title{Differentially Private Video Activity Recognition}

\author{
Zelun Luo$^{1}$\;
Yuliang Zou$^{2}$\;
Yijin Yang$^{3}$\;
Zane Durante$^{1}$\;
De-An Huang$^{4}$
\\
Zhiding Yu$^{4}$\;
Chaowei Xiao$^{3,4}$\;
Li Fei-Fei$^{1}$\;
Animashree Anandkumar$^{4,5}$
\\
\\
{
$^{1}$Stanford University\;
$^{2}$Virginia Tech\; 
$^{3}$Arizona State University\;
$^{4}$NVIDIA\;
$^{5}$Caltech\;
}
}

\maketitle

\begin{abstract}

In recent years, differential privacy has seen significant advancements in image classification; however, its application to video activity recognition remains under-explored. This paper addresses the challenges of applying differential privacy to video activity recognition, which primarily stem from: (1) a discrepancy between the desired privacy level for entire videos and the nature of input data processed by contemporary video architectures, which are typically short, segmented clips; and (2) the complexity and sheer size of video datasets relative to those in image classification, which render traditional differential privacy methods inadequate. To tackle these issues, we propose Multi-Clip DP-SGD, a novel framework for enforcing video-level differential privacy through clip-based classification models. This method samples multiple clips from each video, averages their gradients, and applies gradient clipping in DP-SGD without incurring additional privacy loss. Moreover, we incorporate a parameter-efficient transfer learning strategy to make the model scalable for large-scale video datasets. Through extensive evaluations on the UCF-101 and HMDB-51 datasets, our approach exhibits impressive performance, achieving $81\%$ accuracy with a privacy budget of $\epsilon=5$ on UCF-101, marking a $76\%$ improvement compared to a direct application of DP-SGD. Furthermore, we demonstrate that our transfer learning strategy is versatile and can enhance differentially private image classification across an array of datasets including CheXpert, ImageNet, CIFAR-10, and CIFAR-100. 

\end{abstract}

\begin{table*}[t]
\centering

\resizebox{0.85\linewidth}{!}{
\begin{tabular}{@{}llcccccc@{}}
 \toprule
& Method & Architecture(s) & Pre-training & MNIST & F-MNIST & CIFAR-10 & ImageNet\\
 \midrule
\parbox[t]{20mm}{\multirow{8}{*}{{w/o pre-train}}}
& DP-SGD \cite{abadi2016deep}                   & 2-layer NN & - & 95.00@2.00 & -          & -          & - \\
& DPNAS \cite{cheng2021dpnas}                   & DPNASNet & - & 98.57@3.00 & 88.09@3.00 & 68.33@3.00 & - \\
& Tempered Sigmoid \cite{papernot2020tempered}  & 6/6/9-layer CNN & - & 98.10@2.93 & 86.10@2.70 & 66.20@7.53 & - \\
& DP-ScatterNet \cite{tramer2020differentially} & ScatterNet & - & 98.70@2.93 & 89.70@3.00 & 69.30@3.00 & - \\
& Norm-DP-SGD \cite{davody2020effect}           & LeNet-5/VGG-16 & - & 98.18@3.00 & -          & 77.40@2.00 & - \\
& Private-kNN \cite{Zhu_2020_CVPR}              & 5-layer CNN & - & 98.80@0.47 & - & -          & - \\
& AdaCliP \cite{pichapati2019adaclip}           & 2-layer NN & - & 95.56@2.00 & -          & -          & - \\
& DDP-SGD \cite{du2021dynamic}.                 & 6-layer CNN & - & 96.34@1.20 & 83.81@2.00          & -          & - \\
 \midrule
 \parbox[t]{20mm}{\multirow{5}{*}{{w/ pre-train}}}
& DP-SGD \cite{abadi2016deep}                   & 4-layer CNN & CIFAR-100 & -          & -          & 67.00@2.00 & - \\
& Private-kNN \cite{Zhu_2020_CVPR}              & 4-layer CNN & CIFAR-100 & - & - & 70.80@2.92          & - \\
& Scalable \cite{luo2021scalable}               & ResNet-18 & ImageNet & - & - & 81.57@1.50          & - \\
& DP-ScatterNet \cite{tramer2020differentially} &  ResNet-50 & ImageNet & -          & -          & 92.70@2.00 & - \\
& DP-ImageNet \cite{kurakin2022toward}          & ResNet-18 & Places365 & -          & -          & -          & 47.90@10.00 \\
 \bottomrule
 
\end{tabular}
}
\vspace{-0.2cm}
\caption{\tb{State-of-the-Art Differential Privacy Methods.} Evaluation of differential privacy methods on diverse vision datasets using the acc@$\epsilon$ metric, selecting results nearest to $\epsilon = 3$ for each method. The table emphasizes the limited exploration of large-scale vision tasks and privacy-sensitive domains in current literature.
}

\label{tab:sota}
\vspace{-0.4cm}
\end{table*}

\section{Introduction}
\label{sec:intro}

The rising adoption of machine learning in privacy-sensitive sectors has intensified the need for privacy-preserving machine learning~\cite{davenport2019potential,xiao2018iot,yeung2019computer}. Within these realms, videos constitute a substantial portion of the data, such as surveillance streams used for patient monitoring~\cite{Gerke_2020}. This underscores the criticality of developing video models that are robust against privacy attacks inherent to machine learning~\cite{Papernot_2017,Salem_2019,Shokri_2017}. Notable advancements in image classification models have been achieved through the employment of differential privacy~\cite{abadi2016deep,de2022unlocking,kurakin2022toward,mehta2022large,papernot2018scalable}, furnishing probabilistic privacy assurances by minimizing the impact of data point substitutions within the dataset. However, the field is yet to witness equivalent breakthroughs in video classification, leaving uncertainties regarding the extent to which the accomplishments in image classification can be extended to videos.

We pinpoint two characteristics inherent to video classification that hinder the straightforward adaptation of differential privacy to videos. Firstly, existing private image classification models primarily assume \emph{per-sample} differential privacy, offering probabilistic guarantees for each individual sample (\ie image, video clip) fed into the model during training. This becomes an issue for state-of-the-art video classification models that usually process multiple short clips instead of full videos. Under per-sample differential privacy, these models only assure privacy for the short clips used in training, not for the entire videos. This poses a challenge for state-of-the-art video classification models~\cite{feichtenhofer2019slowfast,liu2022video} which, in various training iterations, sample multiple short \emph{clips} from the same video instead of processing the entire video as a single sample. Under per-sample differential privacy, privacy is assured only for these individual short clips, and not for the whole video from which they are sampled. Secondly, as illustrated in~\tabref{sota}, video datasets that are commonly used, such as UCF-101~\cite{soomro2012ucf101}, HMDB-51~\cite{kuehne2011hmdb}, and Kinetics~\cite{carreira2017quo}, are substantially larger in scale compared to the image classification datasets like MNIST and CIFAR-10, which are traditionally employed in differential privacy research~\cite{papernot2018scalable,abadi2016deep}. Directly employing differential privacy algorithms like DP-SGD on these larger video datasets is impractical due to its detrimental impact on large-scale training.

\topic{Summary of Contributions:} (1) We achieve the first substantial result on differentially private video action classification ($+76\%@\epsilon=5$ on UCF-101) by simultaneously addressing the two aforementioned challenges.
(2) We establish a framework for video-level differential privacy and introduce a novel multi-clip method (Multi-Clip DP-SGD) that allows clip-based video classification models to attain video-level differential privacy without demanding additional privacy budgets.
(3) Through an exhaustive analysis, we pinpoint the most effective parameter-efficient transfer learning strategy for the application of differentially private training to large-scale video datasets.

\begin{figure*}[!ht]
\centering
\includegraphics[width=0.9\textwidth]{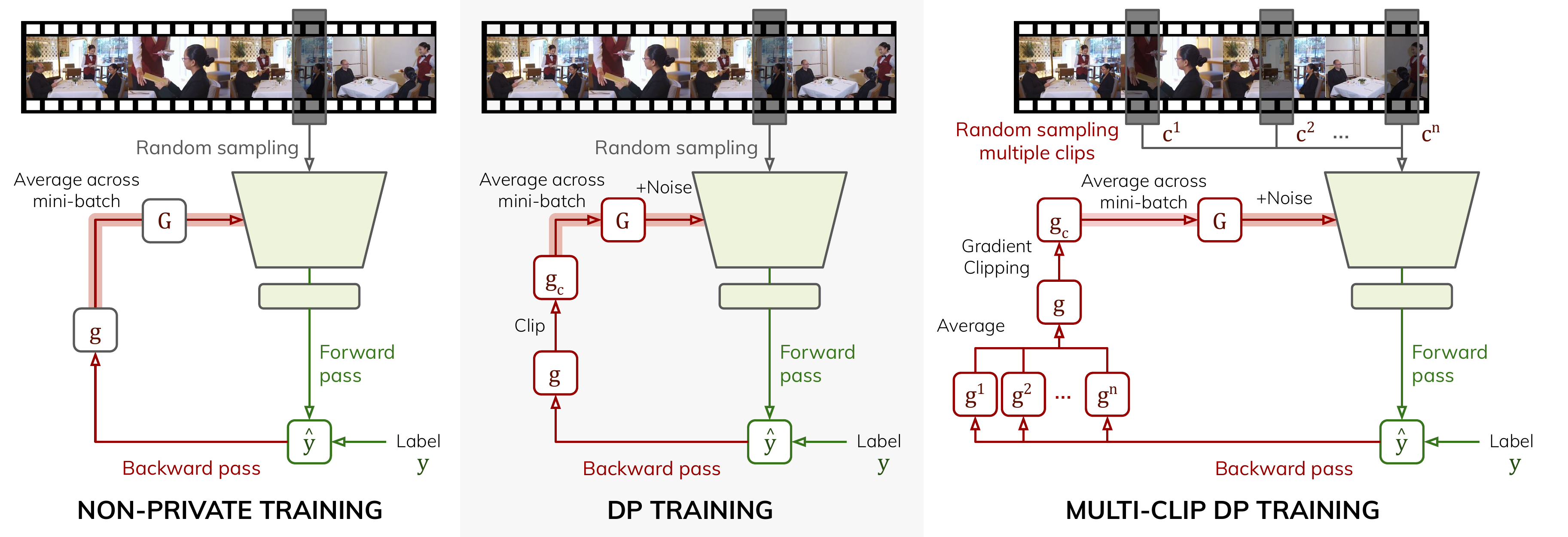}
\vspace{-0.3cm}
\caption{
\tb{Comparison of Training Methods for Videos.} We compare non-private training (left), DP-SGD (middle), and Multi-Clip DP-SGD (right). DP-SGD utilizes gradient clipping and adds Gaussian noise to gradients per video clip. Multi-Clip DP-SGD selects several clips per iteration and combines gradients within a video prior to clipping. By ensuring that each video is observed only once per training step, the performance is enhanced without compromising privacy.
}

\vspace{-0.2cm}
\label{fig:multiclip}
\end{figure*}

Firstly, we recognize a disparity between the level of privacy sought (encompassing an entire video) and the nature of inputs processed by modern video classification architectures~\cite{feichtenhofer2019slowfast,liu2022video} (which are short clips of frames). Conventional training methods focus on clips, and employing differentially private training algorithms like DP-SGD under this setup merely achieves clip-level privacy, compromising the privacy of the full videos. To tackle this issue, we put forward a video-based, multi-clip differential privacy training scheme termed Multi-Clip DP-SGD (\figref{multiclip}). For each selected video, our approach samples several clips per iteration. Rather than directly clipping gradients for each clip, which would inflate the privacy budget in proportion to the number of clips and thus impede the balance between privacy and utility, we average the gradients among clips from the same video before applying the clipping. Our method harnesses the rich information in each video without escalating the privacy budget. 

\begin{figure*}[!ht]
\centering
\includegraphics[width=0.9\textwidth]{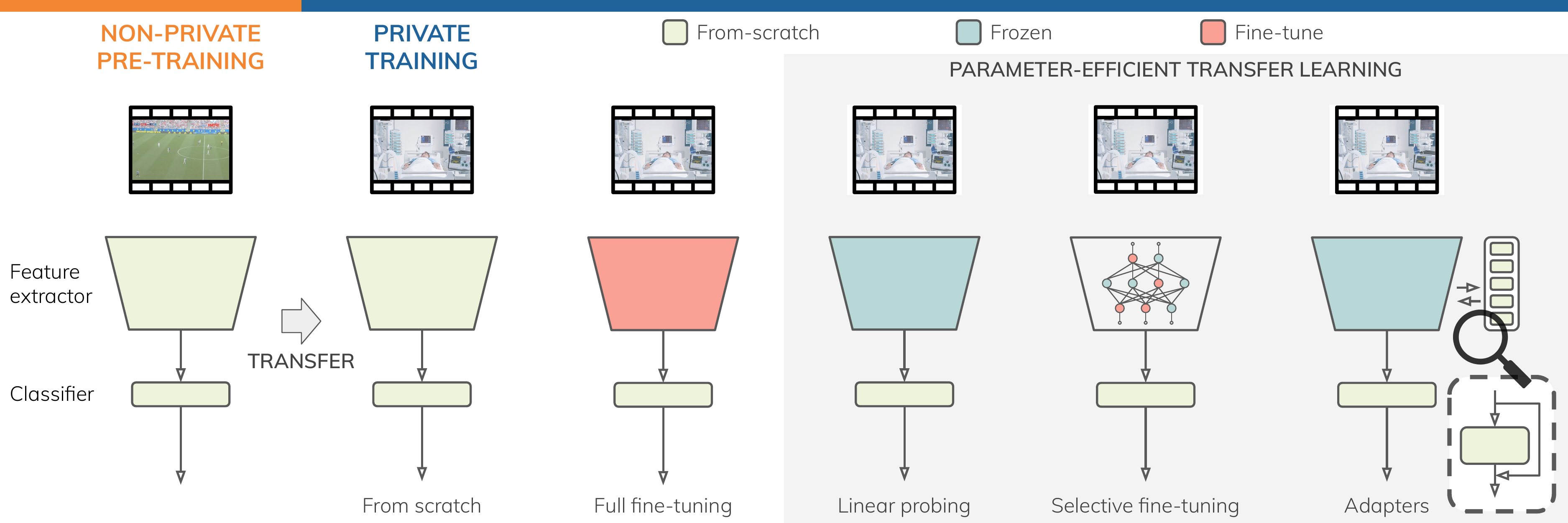}
\vspace{-0.2cm}
\caption{\tb{Transfer Learning Paradigms.} This figure compares four transfer learning paradigms: full fine-tuning, linear probing, selective fine-tuning, and adapters. Notably, the latter three paradigms are characterized as parameter-efficient as they focus on training smaller subnetworks as opposed to the entire network.}
\label{fig:method}
\vspace{-0.4cm}
\end{figure*}

Secondly, we recognize that video classification models and datasets considerably exceed the size of their image classification counterparts typically examined in differential privacy research (UCF-101~\cite{soomro2012ucf101} with 2,916,000 frames compared to MNIST/CIFAR-10 with 60,000 images). Transfer learning has been pivotal in successfully scaling differential privacy to large datasets like ImageNet~\cite{russakovsky2015imagenet} in the realm of image classification~\cite{de2022unlocking,kurakin2022toward,mehta2022large}. By publicly pre-training models on even more extensive datasets like JFT-3B~\cite{sun2017revisiting}, the efficacy of private fine-tuning on target datasets is significantly enhanced. Yet, the optimal fine-tuning strategy, especially for video architectures, remains elusive. To address this, we undertake an exhaustive analysis of transfer learning strategies that are parameter-efficient for differential privacy, as depicted in~\figref{method}. Our study reveals two key insights: (1) fine-tuning normalization layers along with the final linear layer yields the best performance with a negligible increase in trainable parameters; (2) incorporating additional parameters from adapters~\cite{houlsby2019parameter} proves advantageous when there is a substantial domain disparity in pre-training. Leveraging these insights along with our multi-clip video training, we conduct experiments on UCF-101 and HMDB-51 and achieve unprecedented results in video classification. In particular, the LayerNorm~\cite{ba2016layer} architecture (in MViT~\cite{fan2021multiscale}), parameter-efficient transfer learning, and our multi-clip differential privacy training (Multi-Clip DP-SGD) collectively account for our substantial gains ($+76\%@\epsilon=5$ compared to training from scratch).

Finally, we extend our insights to large-scale image classification. Our analysis reveals that contemporary architectures employing LayerNorm~\cite{ba2016layer}, such as ViT-S~\cite{dosovitskiy2020image} and ConvNeXt-T~\cite{liu2022convnet}, outperform the traditionally used ResNet-50-GroupNorm~\cite{kurakin2022toward,wu2018group} in privacy-utility trade-offs, despite comparable parameter counts and ImageNet-1K~\cite{russakovsky2015imagenet} performance. These discoveries propel the state-of-the-art in differential privacy methods, showing substantial gains on standard benchmarks (e.g., $+21.7\%@\epsilon=4$ on CIFAR-10, $+71.1\%@\epsilon=4$ on CIFAR-100) and opening avenues for applying differential privacy to previously unexplored privacy-sensitive datasets like CheXpert~\cite{irvin2019chexpert}.

\section{Related work}
\label{sec:related}

\topic{Differentially-Private SGD (DP-SGD).}
($\epsilon,\delta$)-differential privacy has become the gold standard for database and model privacy due to its probabilistic guarantees of minimal data leakage \cite{dwork_dp_paper,dwork2014algorithmic}.
The most common method for ensuring differential privacy for deep learning models is DP-SGD proposed in Abadi~\etal~\cite{abadi2016deep}.  This modification on standard stochastic gradient descent (SGD) uses gradient clipping and noise additions to mitigate the effects of individual data points \cite{abadi2016deep}. 
There have been several follow-up works \cite{dormann2021not,ghazi2021deep,wang2021dplis} on DP-SGD.  Some explore adaptive methods for clipping gradients to help learning~\cite{pichapati2019adaclip}.  Others propose methods for loss function smoothing to mitigate the effects of the noise added to the gradient~\cite{wang2021dplis}.  Our work uses the accounting method based on Rényi differential privacy \cite{Mironov17Renyi}, which provides a tighter bound on the privacy-utility estimate than the accounting method used in Abadi~\etal~\cite{abadi2016deep}.  
There have also been a series of works exploring the effect of neural architecture choices on the performance obtained when using DP-SGD \cite{morsbach2021architecture,davody2020effect,papernot2020tempered}.  

\topic{Large-Scale Differential Privacy.}
Recent works have shown that differential privacy is possible even on large-scale datasets like ImageNet by leveraging transfer learning, but requires significant pre-training \cite{de2022unlocking,kurakin2022toward}.  Prior work shows that using regular DP-SGD to train on ImageNet achieves an accuracy of $0-1\%$ \cite{kurakin2022toward}.  However, pre-training image models on large-scale internet data has led to reasonable performances on ImageNet, with Kurakin~\etal~\cite{kurakin2022toward} getting $47.9\%$ image classification accuracy with $\epsilon=10$, $\delta = 8 \cdot 10^{-7}$.  In natural language processing (NLP), recent work has shown that large pre-trained language models can be effective differentially private learners \cite{anil2021large,li2021large}.
 
\topic{Transfer Learning.}
In \textit{transfer learning}, a model is first pre-trained on an initial dataset and then fine-tuned on a downstream dataset \cite{pan2009survey,zhuang2020comprehensive}.  
There are many ways to fine-tune models.  The simplest way is full fine-tuning, which trains the entire model on the downstream dataset \cite{kumar2022fine}.  Another way to fine-tune is to freeze the entire model except the last layer and use it as a feature extractor, and then train a linear probe classifier using the feature extractor embedding space \cite{chen2020simple,galanti2021role}.  There have also been ongoing research threads using parameter efficient fine-tuning in order to save computation costs. Prompt-tuning has recently become a common strategy in natural language processing (NLP) for efficient fine-tuning of large language models \cite{lester2021power,liu2021pre}, and adapter modules have also recently become a popular way to efficiently fine-tune large models \cite{gao2021clip,he2021effectiveness}. In alternative approaches to parameter efficient transfer learning, the normalization layers or weights discovered through the lottery ticket hypothesis~\cite{frankle2018lottery} are fine-tuned to enable efficient transfer learning~\cite{li2016revisiting,mehta2019sparse}.


\topic{Video and Privacy.} As videos may contain personal information, there have been several works on privacy-preserving video processing framework. Possible methods include anonymizing faces in videos~\cite{ren2018learning}, cryptographic approach~\cite{pentyala2021privacy}, decreasing the video resolution~\cite{ryoo2017privacy}, and learning anonymization~\cite{wu2020privacy}. While previous research has explored the application of differential privacy to videos, its primary emphasis lies in directly introducing differential privacy to the entire video~\cite{wang2020videodp}. We propose a multi-clip differential privacy training algorithm for video classification, and achieve the first non-trivial result on commonly used video classification datasets.

\newtheorem{definition}{Definition}

\section{Differentially Private Video Classification}
\label{sec:method}

We pinpoint two challenges in differentially private video classification: (1) a discrepancy between model inputs (clips) and privacy objectives (videos); (2) adapting differential privacy for large-scale video datasets. In \secref{pre}, we lay the groundwork with an overview of differential privacy (DP). In \secref{problem}, we delineate video-level DP, the crux of this study. In \secref{multi}, we introduce our solution, Multi-Clip DP-SGD, for video-level DP. In \secref{training}, we outline parameter-efficient transfer learning strategies for scaling differential privacy to videos. Lastly, we examine suitable network architectures in \secref{arch}.

\subsection{Preliminary}
\label{sec:pre}
\noindent\textbf{Differential Privacy.}
Differential privacy~\cite{dwork_dp_paper} provides a formal privacy guarantee to prevent information leakage of each individual data point within a dataset.
By adding randomization to computation over a dataset (\eg training a ML model), the influence of each data point is bounded, and thus the instance-level privacy is protected.
There are two important parameters controlling the strength of differential privacy guarantee: $\epsilon>0$ and $\delta\in[0, 1]$.
And the privacy guarantee becomes stronger as both parameters get smaller.
Formally, we have the following definition, which makes use of the concept of \textit{adjacent datasets} (\ie two datasets than only differ by one data entry).
\vspace{-0.15cm}
\begin{definition}
  \label{def:dp}
  Given a random mechanism $M: D \rightarrow R$ and two adjacent datasets, $d_1, d_2 \in D$, then we say that $M$ satisfies $(\epsilon,\delta)$-differential privacy if $\forall S \subseteq R$:
  \begin{gather*}
  P[M(d_1) \in S] \leq e^{\epsilon} P[M(d_2) \in S] + \delta.
  \end{gather*}
\end{definition}
\vspace{-0.15cm}

\topic{Differentially Private SGD (DP-SGD).}
DP-SGD~\cite{abadi2016deep} is one of the most common strategies to protect the privacy of training data.
With slight modifications upon the traditional SGD algorithm, DP-SGD effectively provides instance-level privacy protection.
More specifically, it has the following modifications.
First, instead of computing mini-batch level gradients at each iteration, DP-SGD computes the per-sample gradient and clips gradient values greater than a constant $C$.
Then, these clipped gradient values are averaged over the training mini-batch and Gaussian noises $\mathcal{N}(0, \sigma^2C^2\mathbf{I})$ are added. DP-SGD was designed to integrate seamlessly with deep neural networks due to its simplicity of implementation and no additional assumptions about the data. \figref{multiclip} compares training with and without DP-SGD for a clip-based video classification model.

\topic{Privacy Accountant.}
A privacy accountant keeps track of the privacy spendings during model training.
The momentum accountant~\cite{abadi2016deep} was first proposed to compute the privacy loss for DP-SGD.
Later, R\'enyi differential privacy was proposed to relax
differential privacy by using the R\'enyi divergence \cite{Mironov17Renyi}. It provides a tighter privacy-utility bound and thus can allow for greater accuracy with the same probabilistic guarantees as standard DP-SGD.

\subsection{Video-Level Differential Privacy}
\label{sec:problem}

In this work, our goal is differentially private video classification. We aim to train a video classifier $f(\cdot)$, which takes a video $v_i$ as input and outputs its class. In addition, $f(\cdot)$ should be trained with privacy guarantees. More specifically, we are interested in training $f(\cdot)$ with \emph{video-level} differential privacy. Recall that \defref{dp} builds on the concept of adjacent datasets. By video-level DP, we mean that each data entry in the adjacent datasets is a video, and the two datasets only differ by one video.

While our goal is to classify the entire video, modern video classification architectures are often designed to take multiple \emph{clips} consisting of several frames as input during training~\cite{fan2021multiscale,liu2022video}.
Each video can be temporally chunk into $N_i$ clips: $v_i = [c_i^1 \dots c_i^{N_i}]$. For simplicity, we assume that the clips are consecutive and non-overlapping, and by concatenating all the $c_i^j$ by order we have the original video $v_i$. Clip-based architecture is a popular and effective design for video classification. However, this creates problems for applying differential privacy to video classification.

Since the video models are trained on clips, if we directly apply privacy-preserving  algorithms in training, such as DP-SGD, then each data entry in the adjacent datasets of \defref{dp} is no longer a video, but instead a clip. In this case, the privacy gaurantees are given for clips instead of videos. We refer to this as \emph{clip-level} differential privacy. It is important to note that clip-level differential privacy does not imply video-level differential privacy for the same $(\epsilon,\delta)$. 
Video-level differential privacy considers adjacent datasets differ by one video $v_i$, which are in fact $N_i$ clips. This is beyond what is covered by clip-level differential privacy (\ie adjacent datasets differ by only one clip).
Since a video $v_i = [c_i^1 \dots c_i^{N_i}]$ consists of a group of clips $c_i^j$, one possible approach is to apply differential privacy's property of \emph{group privacy} to translate clip-level differential privacy to video-level differential privacy~\cite{dwork_dp_paper}.  However, this would simply multiply the privacy budget by the number of clips in videos, which significantly harms the privacy-utility tradeoff. 

\subsection{Multi-Clip DP-SGD for Videos}
\label{sec:multi}

We have discussed that naively applying privacy-preserving training algorithms to video model training only achieves clip-level DP, which does not imply our desired video-level DP. Now we discuss our solution, Multi-Clip DP-SGD, which builds on DP-SGD~\cite{abadi2016deep} to fully exploit information in videos during training, while not increasing their sensitivity in DP.

The first step is to sample by videos instead of clips in training. 
Next, one naive way to process the sampled video is to sample only one clip per video in each mini-batch during the training time, since the architectures are designed to take a clip as input. However, this severely wastes the privacy budget, since we only utilize a small portion of information but with the price of increased privacy loss on the whole video. Repeating this process in training does not help with the privacy-utility tradeoff.



Inspired by the user-level privacy protection~\cite{mcmahan2017learning}, we propose a new multi-clip scheme for video-level differential privacy training. The core idea  is to increase the usage of each video, while not increasing the sensitivity/influence of each video. 
\figref{multiclip} illustrates our Multi-Clip DP-SGD for videos.
First, we sample multiple clips from a video, and compute the per-sample gradient for each clip as in normal DP-SGD. However, instead of clipping the gradient for each clip, we first \emph{average} the gradients from all the clips before clipping the gradient (and adding Gaussian noise). 


Since there is no information exchange between different videos, the computation above satisfies video-level DP.
And since we average multiple clips from the same video before gradient clipping, we can regard that each video is seen only once in each forward pass.
And thus the privacy loss does not increase. As we will show in the experiments, Multi-Clip DP-SGD better leverages the information in each video without incurring additional privacy costs and improves the privacy-utility tradeoff of video classification models.

%

\subsection{Parameter-Efficient Transfer Learning}
\label{sec:training}

While we have proposed Multi-Clip DP-SGD for videos, by itself it is still not sufficient to apply differential privacy to commonly used video datasets and models because of their scales. Even comparatively smaller video datasets like UCF-101~\cite{soomro2012ucf101} and HMDB-51~\cite{kuehne2011hmdb} are still much larger than CIFAR datasets used in differential privacy literature. 
Recent works have shown that transfer learning provides a significant benefit to DP-SGD in terms of privacy-utility trade-off and scaling to large image classification tasks~\cite{tramer2020differentially,li2021large}. 
Building on this strategy, we identify \emph{parameter-efficient} transfer learning as a promising direction for scaling differential privacy to video datasets and conduct a comprehensive study along this direction.

Specifically, DP-SGD aims to minimize the influence of the training data during
the training process by (1) clipping the $\ell_2$ norm of per-sample gradient $\mathbf{g}$ at $\mathbf{g}/\max(1,\lVert\mathbf{g}\rVert_2/C)$, and (2) adding Gaussian noise sampled from $\mathcal{N}(0, \sigma^2C^2\mathbf{I})$, where $\sigma$ and $C$ are gradient norm bound and noise scale respectively. The above analysis indicates that DP-SGD favors models with smaller number of trainable parameters: the isotropic Gaussian noise applied to training gradients has a greater impact on models with more trainable parameters because the expected norm of the noise increases with the number of parameters~\cite{li2021large}. 
On the other hand, modern deep learning models tend to be highly overparameterized~\cite{he2016deep,krizhevsky2012imagenet} and researchers have shown that increasing the number of parameters leads to improved generalization~\cite{neyshabur2018towards}. Parameter-efficient transfer learning provides an opportunity to use these large models in DP, while not increasing the number of trainable parameters. Next, we discuss the (parameter-efficent) transfer learning schemes we study in this paper.



\topic{Full Fine-Tuning.}
All model parameters are trained on the target data for full fine-tuning, which makes it not parameter-efficient. We select this approach to compare with other parameter-efficient approaches. 
Although full fine-tuning is probably the most common approach in non-DP transfer learning scenarios, it usually does not work well under differential privacy settings.
Differential privacy struggles with more trainable parameters since DP-SGD (clipping/noise) negatively affects models with more trainable parameters.

\topic{Linear Probing.}
In this setting, all model parameters are fixed except the last linear layer.
The frozen part can be regarded as a feature extractor, and we train a liner classification on top of the extracted features.
Since the last linear layer only contains a small portion of parameters (compared to the full model), DP-SGD can be well combined with linear probing to provide a decent privacy guarantee.
However, the feature extractor is trained on the source domain, and thus training only a linear layer may be sub-optimal due to the existence of domain shifts.

\topic{Selective Fine-Tuning.}
As a trade-off between full fine-tuning and linear probing, one can choose to train a small portion of parameters in the feature extractor, in addition to the last linear layer.
This strategy allows a model to adapt itself to the target distribution while still maintaining a reasonable amount of trainable parameters.
And thus, DP-SGD can be effectively applied under such a setting to provide a good privacy-utility trade-off. Previous work has used Lottery Ticket Hypothesis~\cite{frankle2018lottery} to select the parameters to train~\cite{luo2021scalable}.
In this paper, we choose to train the normalization layers in addition to the last linear layer.

\topic{Adapter Training.}
Similar to selective fine-tuning, adapter training also aims to train a small number of model parameters in addition to the last linear layer.
However, instead of updating the parameters in the original model architecture, we incorporate additional parameters inside the current network and train them  together with the last linear layer.
Similar to adapter training in non-DP settings~\cite{gao2021clip,houlsby2019parameter,zhang2021tip}, we design lightweight 2-layer MLPs with skip-connection and insert these modules inside the pre-trained networks.
These modules are initialized as identity mappings and gradually trained to stabilize model training.

\subsection{Architecture}
\label{sec:arch}

Finally, we further investigate model architectures that are suitable for training with DP. In particular, we find that architectures with different types of normalization layers are worth further investigation.
BatchNorm~\cite{ioffe2015batch} is empirically proved to stabilize model training and speed up convergence~\cite{wu2021rethinking}, and thus it has been an essential component in most modern CNNs.
However, BatchNorm layers are not supported in differentially-private models.
The training time behavior of BatchNorm requires computation of mean and variance of each input mini-batch, creating a dependency between samples, which violates the formulation of differential privacy.
There are two ways to resolve this issue.
The first one is to replace BatchNorm with other normalization layers (\eg GroupNorm~\cite{wu2018group}, LayerNorm~\cite{ba2016layer}). This solution seems to be sub-optimal, since the new normalization layers are not designed for these CNNs.
The second option is to use models without BatchNorm by its original design. We find that the recent vision transformers are good fits in this case.  We further study a new CNN architecture, ConvNeXt. 
In this paper, we study two different types of network architectures: convolutional neural networks (CNN) and vision transformers.
For CNNs, we adopt ResNet~\cite{he2016deep} and ConvNeXt~\cite{liu2022convnet}.
For vision transformers, we adopt ViT~\cite{dosovitskiy2020image} in image experiments and MViT~\cite{fan2021multiscale} in video experiments.

\begin{table}[t]
\centering
\resizebox{0.85\linewidth}{!}{
\begin{tabular}{@{}lccccccccc@{}}
\toprule
Scheme
& \#Params.
& \#Clips
& $\epsilon=5$
& $\epsilon=10$
\\
\midrule

From scratch & 36.4M & 1 & 4.93 & 6.60 \\

Full fine-tune & 36.4M & 1 & 29.15 & 38.58 \\

Linear probe & 77.7K & 1 & 75.05 & 78.17 \\

Linear probe & 77.7K & 8 & 77.96 & 80.49 \\

Adapter & 1.8M & 8 & 79.19 & 81.73 \\

Selective fine-tune & 110K & 8 & \textbf{80.86} & \textbf{82.81} \\

\bottomrule
\end{tabular}
}
\vspace{-0.2cm}
\caption{\tb{Experiments on UCF-101.}
We adopt $\delta=10^{-5}$ and report the top-1 accuracy.
We conduct experiments using an MViT-B/16$\times$4 model pre-trained on the Kinetics-400 dataset.
}
\vspace{-0.4cm}
\label{tab:ucf}
\end{table}

\begin{table}[t]
\centering

\resizebox{0.85\linewidth}{!}{
\begin{tabular}{@{}lccccccccc@{}}
\toprule
Scheme
& \#Params.
& \#Clips
& $\epsilon=5$
& $\epsilon=10$
\\

\midrule

From scratch & 36.4M & 1 & 3.62 & 4.13 \\

Full fine-tune & 36.4M & 1 & 17.93 & 21.21 \\

Adapter & 1.8M & 1 & 50.34 & 53.37 \\

Selective fine-tune & 110K & 1 & 51.68 &  56.48  \\

\midrule

From scratch & 36.4M & 8 & 6.87 & 7.24 \\

Full fine-tune & 36.4M & 8 & 29.20 & 33.75 \\

Adapter & 1.8M & 8 & 53.62 & 56.23 \\

Selective fine-tune & 110K & 8 & \textbf{57.49} & \textbf{60.52} \\

\bottomrule
\end{tabular}
}
\vspace{-0.2cm}
\caption{\tb{Experiments on HMDB-51.}
We adopt $\delta=10^{-5}$ and report the top-1 accuracy.
We conduct experiments using an MViT-B/16$\times$4 model pre-trained on the Kinetics-400 dataset.
}
\vspace{-0.4cm}
\label{tab:hmdb}
\end{table}


\begin{figure*}[!ht]
    \begin{minipage}[t]{.32\textwidth}
        \centering
        \includegraphics[width=\textwidth]{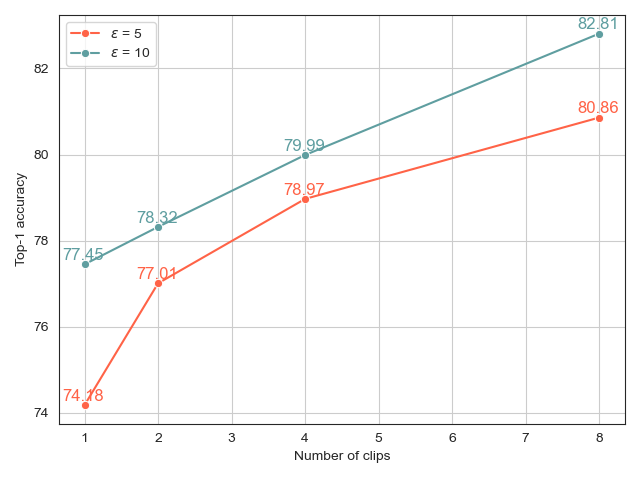}
        \subcaption{UCF-101}
        \label{fig:ucf}
    \end{minipage}
    \hfill
    \begin{minipage}[t]{.32\textwidth}
        \centering
        \includegraphics[width=\textwidth]{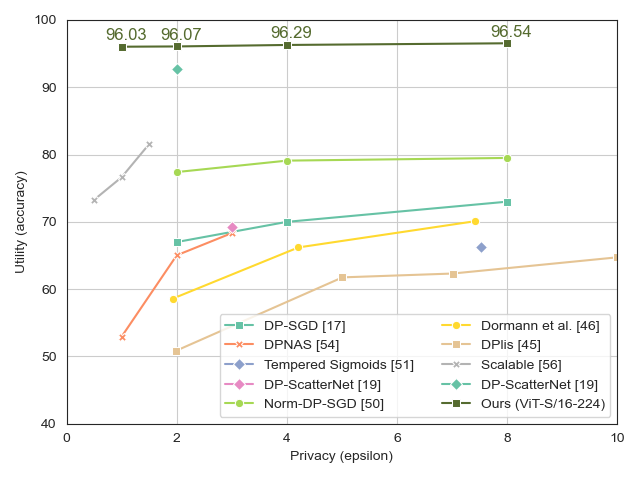}
        \subcaption{CIFAR-10}
        \label{fig:cifar10}
    \end{minipage}
    \hfill
    \begin{minipage}[t]{.32\textwidth}
        \centering
        \includegraphics[width=\textwidth]{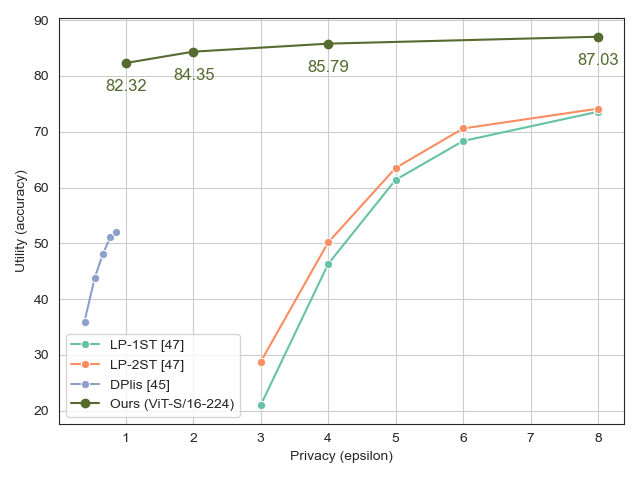}
        \subcaption{CIFAR-100}
        \label{fig:cifar100}
    \end{minipage}
    \vspace{-0.2cm}
    \caption{\tb{Quantitative Analysis on UCF-101, CIFAR-10, and CIFAR-100.} (a) evaluates Multi-Clip DP-SGD for video classification on UCF-101 by experimenting with varying numbers of clips ($1, 2, 4, 8$) and reporting the top-1 accuracy. (b) and (c) offers a comparative analysis on CIFAR-10 and CIFAR-100 against state-of-the-art differential privacy methods, reporting top-1 accuracy with $\delta=10^{-5}$.}
    \label{fig:sota}
    \vspace{-0.1cm}
\end{figure*}

\begin{table*}[htbp]
\centering
\setlength{\tabcolsep}{10pt}

\resizebox{0.75\linewidth}{!}{
\begin{tabular}{@{}llcccccccccc@{}}
\toprule
Architecture
& Scheme
& \#Trainable Params.
& $\epsilon=1$
& $\epsilon=2$
& $\epsilon=4$
& $\epsilon=8$
& $\epsilon=\infty$
\\
\midrule
\parbox[t]{20mm}{\multirow{4}{*}{{ResNet-50-GN}}}
& From scratch & 23.7M & 4.16 & 7.29 & 11.83 & 17.25 & -\\
& Full fine-tune & 23.7M & 52.96 & 59.63 & 66.92 & 71.32 & 82.40 \\
& Linear probe & 204K & 62.14 & 65.89 & 68.71 & 70.28 & - \\
& Selective fine-tune & 258K & \textbf{62.33} & \textbf{68.93} & \textbf{72.32} & \textbf{74.69} & -\\

\midrule
\parbox[t]{20mm}{\multirow{5}{*}{{ViT-S/16-224}}}
& From scratch & 21.7M & 13.76 %
& 16.22 & 19.67 & 22.43 & - \\
& Full fine-tune & 21.7M & 79.36 & 83.64 & 85.67 & 86.59 & 90.27 \\
& Linear probe & 38.5K & 71.64 & 74.81 & 76.47 & 77.03 & -   \\
& Selective fine-tune & 57.7K & \textbf{84.73} & \textbf{86.90} & \textbf{87.68} & \textbf{88.43} & - \\
& Adapter & 928K & 82.32 & 84.35 & 85.79 & 87.03 & -  \\

\midrule
\parbox[t]{20mm}{\multirow{5}{*}{{ConvNeXt-T}}}
& From scratch & 27.9M & 9.20 %
& 12.93 & 15.97 & 18.45 & - \\
& Full fine-tune   & 27.9M & 65.93 & 74.00 & 79.17 & 81.22 & 89.98 \\
& Linear probe     & 76.9K & 76.59 & 78.82 & 80.32 & 80.91 & - \\
& Selective fine-tune & 93.2K & \textbf{83.11} & \textbf{84.69} & \textbf{86.06} & \textbf{86.67} & - \\
& Adapter          & 1.7M & 79.37 & 83.68 & 85.11 & 86.05 & - \\

\bottomrule
\end{tabular}
}

\vspace{-0.2cm}
\caption{\tb{Ablation Study on CIFAR-100.} The models are trained for 50 epochs with $\delta=10^{-5}$, and the top-1 accuracy is reported to validate the effectiveness of the design choices.}

\vspace{-0.2cm}
\label{tab:ablation}
\end{table*}

\section{Experiments}
\label{sec:results}

Now we evaluate our proposed Multi-Clip DP-SGD and parameter-efficient transfer learning for video classification. In addition, we further evaluate our parameter-efficient transfer learning on scaling image classification and compare it with state-of-the-art approaches in DP.

\subsection{Differentially Private Video Classification}

\topic{Experimental Setup.} We use UCF-101~\cite{soomro2012ucf101} and HMDB-51~\cite{kuehne2011hmdb} for our evaluation of differentially private video classification.
UCF-101 is a widely used action recognition dataset that contains 133,20 action videos annotated into 101 action classes.
HMDB-51 is a video classification dataset of human motion,
containing 6849 clips divided into 51 action categories, each containing a minimum of 101 clips.
Unlike image classification, video classification~\cite{simonyan2014two,feichtenhofer2019slowfast} of human activities typically only samples a clip of frames from each video in a training batch. We report top-1 accuracy as the utility metric. To the best of our knowledge, this is the first study of differentially private video classification on a large-scale video dataset. 

\topic{Parameter-Efficient Transfer Learning.} The results are shown in \tabref{ucf} and \tabref{hmdb}. We find that full fine-tuning significantly improves training from scratch (+24\%@$\epsilon=5$ in UCF-101 and +14\%@$\epsilon=5$ in HMDB-51). In addition, linear probing further improves full fine-tuning by limiting the number of trainable parameters (+46\%@$\epsilon=5$ in UCF-101 and +32\%@$\epsilon=5$ in HMDB-51). Among different parameter-efficient transfer learning schemes, selective fine-tuning improves linear probing by 2.9\%@$\epsilon=5$ on UCF-101 with a slight increase of trainable parameters. Adapter's improvement is less than selective fine-tuning in this scenario. As we will see in later on image classification, the relative performance differences of these parameter-efficient schemes (linear, adapter, selective) are dataset dependent. For transfer learning from Kinetics to UCF-101 and HMDB-51, linear probing is already effective because of the small domain gap.


\topic{Multi-Clip DP-SGD.} We study the effect of our Multi-Clip DP-SGD in \figref{ucf}, where we compare results with varying numbers of clips. With only 1 clip, the result is reduced to standard DP-SGD. We use selective fine-tuning for this analysis. With the increasing number of clips in Multi-Clip DP-SGD, the utlity continues to improve given a fixed privacy budget. Overall, our multi-clip approach leads to 6.7\%@$\epsilon=5$ compared to standard DP-SGD, which is larger than the difference between different parameter-efficient schemes. This shows the importance of our multi-clip approach for video classification under DP.

\begin{table*}[htbp]

\centering

\resizebox{0.65\linewidth}{!}{
\begin{tabular}{@{}lllccc@{}}
\toprule
Architecture
& Pre-train
& Batch size
& Scheme
& \#Trainable Params.
& $\epsilon=10$
\\
\midrule

\parbox[t]{20mm}{\multirow{3}{*}{{ViT-B/16-224}}}
& \parbox[t]{20mm}{\multirow{3}{*}{{CLIP~\cite{radford2021learning}}}}
& \parbox[t]{20mm}{\multirow{3}{*}{{65,536}}}
&Full fine-tune & 86.6M & 66.09 \\

&&&Linear probe & 769K & 79.36  \\

&&&Selective fine-tune & 808K & \textbf{79.86}  \\

\midrule

\parbox[t]{20mm}{\multirow{4}{*}{{ViT-S/16-224}}}
& \parbox[t]{20mm}{\multirow{4}{*}{{Places365~\cite{zhou2017places}}}}
& \parbox[t]{20mm}{\multirow{4}{*}{{1,024}}}
&Full fine-tune & 22.1M & 29.34 \\

&&& Linear probe & 385K & 26.53 \\

&&&Selective fine-tune & 404K & \textbf{40.78}  \\

&&&Adapter & 1.3M & 40.47  \\

\midrule

\parbox[t]{20mm}{\multirow{3}{*}{{ViT-S/16-224}}}
& \parbox[t]{20mm}{\multirow{3}{*}{{Places365~\cite{zhou2017places}}}}
& \parbox[t]{20mm}{\multirow{3}{*}{{65,536}}}
&Full fine-tune & 22.1M & 30.21 \\

&&&Selective fine-tune & 404K & \textbf{45.12}  \\

&&&Adapter & 1.3M & 41.79  \\

\bottomrule
\end{tabular}
}
\vspace{-0.2cm}
\caption{\tb{Large-Scale Differential Privacy Experiments on ImageNet.}
We adopt $\delta=10^{-6}$ and report the top-1 accuracy.
}
\vspace{-0.2cm}
\label{tab:imagenet}
\end{table*}

\begin{table*}[htbp]
\centering
\setlength{\tabcolsep}{8pt}

\resizebox{0.75\linewidth}{!}{
\begin{tabular}{@{}llcccccccccc@{}}
\toprule
Architecture
& Scheme
& \#Trainable Params.
& Pre-train
& $\epsilon=1$
& $\epsilon=2$
& $\epsilon=4$
& $\epsilon=8$
\\

\midrule
\parbox[t]{20mm}{\multirow{5}{*}{{ViT-S/16-224}}}
& From scratch & 21.7M & - & 74.82 & 74.93 & 75.13 & 75.60 \\

& Full fine-tune & 21.7M & IN-21k$\rightarrow$IN-1k & 82.57 & 83.50 & \textbf{84.98} & \textbf{85.78} \\

& Selective fine-tune & 21.1K & IN-21k $\rightarrow$ IN-1k & 81.76 & 83.64 & 83.71 & 84.46 \\

& Adapter & 892K & IN-21k$\rightarrow$IN-1k & \textbf{83.31} & \textbf{83.69} & 83.75 & 84.77  \\

\midrule

\parbox[t]{20mm}{\multirow{4}{*}{{ConvNeXt-T}}}
& From scratch & 27.8M & - & 76.93 & 77.19 & 77.51 & 79.08 \\

& Full fine-tune & 27.8M & IN-21k $\rightarrow$ IN-1k & 85.23 & 85.37 & 86.83 & 87.21 \\

& Selective fine-tune & 20.2K & IN-21k $\rightarrow$ IN-1k & 85.06 & 85.29 & 85.92 & 86.56 \\

& Adapter & 1.7M & IN-21k $\rightarrow$ IN-1k & \textbf{86.77} & \textbf{87.06} & \textbf{87.78} & \textbf{88.05} \\

\bottomrule
\end{tabular}
}
\vspace{-0.2cm}
\caption{\tb{Large-Scale Differential Privacy Experiments on CheXpert.}
We adopt $\delta=10^{-5}$ and report the AUROC.
All models were trained for 10 epochs.
}

\vspace{-0.4cm}
\label{tab:chexpert}
\end{table*}

\subsection{Differentially Private Image Classification}


To underscore the potency of our parameter-efficient transfer learning strategies, we put them to the test on large-scale image classification tasks using two datasets: ImageNet and CheXpert. ImageNet is widely regarded as the benchmark dataset, while CheXpert offers insights into the application of differential privacy in privacy-sensitive scenarios with large domain gaps.

\topic{ImageNet.}
ImageNet~\cite{russakovsky2015imagenet} has been widely used for image classification.  It consists of 1,000 classes of 1,281,167 training images and 50,000 validation images. 
We report results on ImageNet using two ViT \cite{dosovitskiy2020image} archiectures and two different pre-training datasets. As shown in Table \ref{tab:imagenet}, full fine-tuning results in worse performance than the parameter-efficient strategies.  We note that using the pre-training dataset from CLIP \cite{radford2021learning} results in substantial performance gains, highlighting the importance of large-scale pre-training for transfer learning in the differentially private setting.  We significantly bridge the gap with supervised performance, with the ViT-B/16 architecture achieving 85.49\% top-1 accuracy on ImageNet with public training~\cite{dosovitskiy2020image}.

\topic{CheXpert (Medical Image Classification).} CheXpert~\cite{Irvin_2019} is a large-scale dataset of chest X-rays annotated by professional radiologists. It consists of 224,316 chest radiographs of 65,240 patients.
Following the standard setting~\cite{Irvin_2019, yuan2020large}, we report the AUC score as the utility metric on five selected diseases, i.e., Cardiomegaly, Edema, Consolidation, Atelectasis, Pleural Effusion. 
Results are in \tabref{chexpert}. Adapter outperforms selective fine-tuning with the large domain gap in this case. In addition, for the ViT backbone, when we have sufficient privacy budget, DP-SGD is sufficient to fine-tune the full model and full fine-tuning performs the best. However, parameter-efficient schemes still outperform with less privacy budgets ($\epsilon=1,2$). 

\subsection{Comparison with State-of-the-Art Methods}

In this section, we conduct an ablation study and compare our approach with state-of-the-art differential privacy methods on CIFAR-100, as shown in \tabref{ablation}.

\topic{Parameter-Efficient Transfer Learning.} We still find that transfer learning outperforms training from scratch by a large margin on  datasets commonly used in differential privacy literature. 
For CNN models (ResNet, ConvNeXt), linear probing outperforms full fine-tuning when using a small $\epsilon$. 
Selective fine-tuning is consistently the best-performing method on CIFAR-100 despite only having a small increase of trainable parameters compared to linear probing. The improvement is more significant for LayerNorm-based architectures (ViT and ConvNeXt).

\topic{LayerNorm-Based Architectures.} We find that using recent architectures based on LayerNorm leads to better performance gains compared to the commonly used ResNet-50, despite having a similar amount of trainable parameters and similar performance when publicly trained. For ResNet-50, the performances of linear probing and selective fine-tuning are much closer. In contrast, on ViT and ConvNeXt, selective fine-tuning leads to more significant improvement. This suggests that selective fine-tuning might synergize better with LayerNorm compared to GroupNorm in our experiments.

\topic{Comparison with SOTA.} We compare our models with state-of-the-art results on CIFAR-10 and CIFAR-100. 
Results on the more challenging CIFAR-100 are more limited compared to CIFAR-10.
As shown in~\figref{sota}, under our parameter-efficient transfer learning scheme based on the previous study, our method consistently outperforms the existing approaches by a large margin, across all measured $\epsilon$ values and across both datasets.

\subsection{Implementation Details}
\label{sec:setup}



We use the implementation from the Opacus library~\cite{opacus} to apply DP-SGD for model training. We utilize distributed training and model checkpointing~\cite{chen2016training} to increase the batch size.
For video experiments, we use the PyTorchVideo library~\cite{fan2021pytorchvideo}.
All the experiments are conducted on a Google Cloud instance with 8 Nvidia A100 GPUs.
While hyper-parameters play a critical role in differentially private training, they are notoriously difficult to tune. In order to facilitate comparison between methods, we pre-define the training epochs for each dataset. And we fix the clipping norm as $C=1$.
We search for the optimal learning rate on the CIFAR-100 dataset with a fixed $\epsilon=1$.
We perform a grid search over the learning rate between $[10^{-4}, 10^{-2}]$.


\vspace{-0.2cm}
\section{Conclusion and Limitation}
\label{sec:conclusions}
We introduce the pioneering application of differential privacy to video classification on the UCF-101 and HMDB-51 datasets.  Our integrated approach, featuring the innovative Multi-Clip DP-SGD and parameter-efficient transfer learning tailored for modern video architectures under DP, achieves a remarkable 76\% accuracy improvement at $\epsilon=5$ compared to the direct application of DP-SGD. One limitation of our study is the use of ($\epsilon,\delta$)-differential privacy as the sole standard for evaluating privacy. The lack of established criteria for practical $\epsilon$ settings necessitates future research to assess models against pragmatic privacy metrics.

{\small
\bibliographystyle{ieee_fullname}
\bibliography{egbib}
}

\end{document}